# Knowledge-guided Convolutional Networks for Chemical-Disease Relation Extraction

Huiwei Zhou[1,*], Chengkun Lang[1], Zhuang Liu[1], Shixian Ning[1], Yingyu Lin[2] and Lei Du[3]

[1]School of Computer Science and Technology, Dalian University of Technology, Address Chuangxinyuan Building, No.2 Linggong Road, Ganjingzi District, Dalian, Liaoning, 116024, China.

[2]School of Foreign Languages, Dalian University of Technology, Address Arts Building, No.2 Linggong Road, Ganjingzi District, Dalian, Liaoning, 116024, China.

[3]School of Mathematical Sciences, Dalian University of Technology, Address Chuangxinyuan Building, No.2 Linggong Road, Ganjingzi District, Dalian, Liaoning, 116024, China.

*To whom correspondence should be addressed.

Email: zhouhuiwei@dlut.edu.cn, kunkun@mail.dlut.edu.cn, zhuangliu1992@mail.dlut.edu.cn, ningshixian@mail.dlut.edu.cn, lyydut@sina.com, dulei@dlut.edu.cn.

## Abstract

**Background:** Automatic extraction of chemical-disease relations (CDR) from unstructured text is of essential importance for disease treatment and drug development. Meanwhile, biomedical experts have built many highly-structured knowledge bases (KBs), which contain prior knowledge about chemicals and diseases. Prior knowledge provides strong support for CDR extraction. How to make full use of it is worth studying.

**Results:** This paper proposes a novel model called "Knowledge-guided Convolutional Networks (KCN)" to leverage prior knowledge for CDR extraction. The proposed model first learns knowledge representations including entity embeddings and relation embeddings from KBs. Then, entity embeddings are used to control the propagation of context features towards a chemical-disease pair with gated convolutions. After that, relation embeddings are employed to further capture the weighted context features by a shared attention pooling. Finally, the weighted context features containing additional knowledge information are used for CDR extraction. Experiments on the BioCreative V CDR dataset show that the proposed KCN achieves 71.28% F1-score, which outperforms most of the state-of-the-art systems.

**Conclusions:** This paper proposes a novel CDR extraction model KCN to make full use of prior knowledge. Experimental results demonstrate that KCN could effectively integrate prior knowledge and contexts for the performance improvement.

## 1  Background

Chemicals, diseases and their relations play important roles in many areas of biomedical research and health care [1-3]. Because of their critical significance, these relations are curated into knowledge bases (KBs) such as the Comparative Toxicogenomic Database[1] (CTD) [4] by domain experts, continually. However, manual curation of chemical-disease relation (CDR) from the literature is costly and difficult to keep up-to-date. Automatic extraction of CDR from texts has become increasingly important.

To promote the research on CDR extraction, the BioCreative-V community proposes a task of automatically extracting CDR from biomedical literature [5], which contains two specific subtasks: (1) disease named entity recognition and normalization (DNER); (2) chemical-induced diseases (CID) relation extraction. This paper focuses on the CID subtask at both intra- and inter-sentence levels. The intra- and inter-sentence levels refer to a chemical-disease pair in the same sentence and in two different sentences, respectively.

Up to now, many methods have been proposed for the automatic extraction of CDR. These methods could be mainly divided into two categories: feature-based methods [6-10] and neural network-based methods [11-17]. Feature-based methods aim at extracting different kinds of context features. Gu et al. [6] devise various effective linguistic features for CDR extraction. Zhou et al. [7] extract the shortest dependency path (SDP) between chemical entities and disease entities, which provide strong evidence for relation extraction. Although complicated handcrafted features achieve good performance, they are time-consuming and difficult to extend to a new dataset.

In recent years, neural network-based relation extraction methods have achieved significant breakthrough: they can model language more precisely with low-dimensional feature vectors rather than one-hot handcrafted features. Gu et al. [11] employ convolutional neural network (CNN) [18] to learn the context and dependency representations for CDR extraction. Zhou et al. [12] use long short-term memory neural network (LSTM) [19] to generate representations of SDP sequences for CDR extraction. Nguyen et al. [13] incorporate character-based word representations into a standard CNN-based relation extraction model. Neural network-based methods could learn semantic features from context sequences automatically and show promising results for CDR extraction.

Besides the context features mentioned above, prior knowledge on chemicals and diseases is also important for relation extraction. Comparative Toxicogenomic Database (CTD) [4] is a well-known biomedical knowledge base, which contains a large amount of structured triples in the form of (*head entity*, *relation*, *tail entity*). Feature-based methods use knowledge features (relations of chemical-disease pairs in the KBs) to extract CID relations [8-10]. They significantly improve the CDR extraction performance. However, one-hot knowledge features assume that all entities and relations are independent from each other, which does not take the semantic relevance into consideration.

To better model prior knowledge in KBs, some researchers focus on knowledge representation learning, which could learn low-dimensional embeddings for entities and relations [20-22]. TransE [20] is a typical translation-based method. It projects entities and relations into a common embedding space, and regards relations as translations from head entities to tail entities in this space.

Neural network-based methods employ relation embeddings learned from CTD to select important context words [17]. With the help of low-dimensional knowledge representations, Zhou et al. [17] efficiently compute semantic links between contexts and relations in a low-dimensional space, which results in an increase in the CDR extraction performance. However, only relation embeddings are utilized as the guidance in their model. Entity embeddings of chemical-disease pairs are completely ignored. Since humans would like to pay more attention to the focused entities while extracting the relation of the entity pair, entity embeddings are helpful for relation extraction.

Recently, some neural network architectures, such as attention-based memory network [23], attention-based LSTM [24] and gated convolutional neural network (GCNN) [25-27] are proposed to grasp important context information. Among them, GCNN with gated convolution operations can generate target-specific features accurately and efficiently [26].

To make full use of the knowledge representations, this paper proposes a novel model called "Knowledge-guided Convolutional Networks (KCN)" for CDR extraction. First, chemical and disease embeddings are used to control the propagation of context features towards the two focused entities through gated convolution operations, respectively. Then, relation embeddings are employed to further capture the weighted context features through a shared attention pooling. Finally, the weighted context features containing additional knowledge information are used to extract CID relations.

The major contributions of this paper are summarized as follows:
- To make full use of both entity embeddings and relation embeddings, we propose a novel model KCN, which introduces gating operations

into the convolutional layer and the attention mechanism into the pooling layer. The experimental results show its effectiveness in capturing knowledge-related context features for relation extraction.
- Gated convolution networks with entity embeddings could selectively output context features related to the focused entity pairs.

## 2 Methods

This section introduces a CDR extraction approach in four steps: (1) extract the candidate instances at both intra- and inter-sentence levels from the CDR dataset; (2) learn knowledge representations from the CTD knowledge base with TransE model; (3) train the knowledge-guided convolutional networks (KCN) on the candidate instances with the guidance of knowledge representations; (4) merge the extraction results at intra- and inter-sentence levels as the final document level results.

### 2.1 Instance construction

#### 2.1.1 Intra- and inter-sentence level instance construction

The candidate chemical-disease instances are constructed at intra- and inter- sentence level separately. All the chemical-disease pairs that exist in the same sentence are extracted as the intra-sentence level instances without any limitation. For the inter-sentence level instances, we employ the following heuristic rules [11] to remove some negative instances.

(1) In the same document, all the intra-level chemical-disease instances will not be considered as inter-sentence level instances.

(2) A chemical-disease pair will not be taken into consideration if the sentence distance between the chemical and disease is more than 3.

(3) If there are multiple mentions that refer to the same entity, only the chemical-disease pairs existing in the nearest distance are considered as the inter-sentence level instances.

#### 2.1.2 Hypernym filtering

A concept of disease or chemical may be hypernym concept to a more specific one. However, the goal of the CID task is to extract the relations between the most specific diseases and chemicals. Therefore, we remove those instances including hyper-entities which are more general than entities already participating in the instances. Specially, the hypernym relationships between entities are determined by indexing the Medical Subject Headings (MeSH) [28].

#### 2.1.3 Shortest dependency path sequence generation

This paper takes SDP sequences as the inputs for CDR extraction. Take sentence 1 as an example of SDP sequence generation:

Sentence 1: *Seizures were induced by pilocarpine injections in trained and non-trained control groups.*

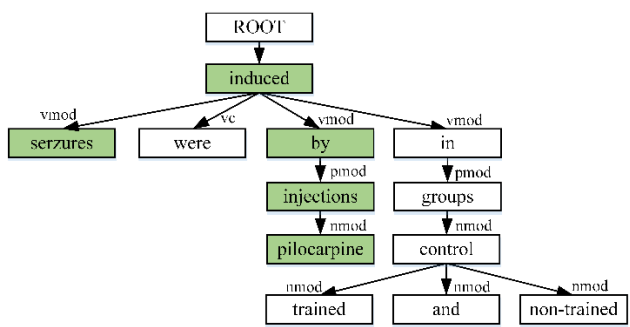

**Fig. 1** The dependency tree of sentence 1 with chemical "*pilocarpine*" and disease "*seizures*".

The chemical entity "*pilocarpine*" is denoted by wave line and the disease entity "*seizures*" is denoted by underline. The corresponding dependency tree is shown in Fig. 1, with the SDP between this entity pair highlighted in green (all the words are transformed to lowercase and the punctuations are discarded). Intuitively, we directly take the SDP sequence {"*pilocarpine*", "↑", "*nmod*", "↑", "*injections*", "↑", "*pmod*", "↑", "*by*", "↑", "*vmod*", "↑", "*induced*", "↓", "*vmod*", "↓", "*seizures*"} as the input of KCN. In this sequence, the symbols "↑" and "↓" indicate the dependency directions, and the tokens like "*vmod*" represent the dependency relation tags between two words. We can find that the trigger word "*induced*" is included in the SDP sequence, which could directly indicate whether the chemical-disease pair has the CID relation, while meaningless words are omitted.

The dependency tree is generated by Gdep Parser [29]. For an intra-sentence level instance, we directly extract the SDP sequence from chemical to disease. For an inter-sentence level instance, we first connect the roots of the dependency trees of the two sentences by using an artificially introduced root. Then, the SDP sequence from the chemical entity to the disease entity is extracted from this new tree.

## 2.2 Knowledge representation learning

This section describes how to use the TransE model to learn knowledge representations based on chemical-disease triples in the form of (*chemical*, *relation*, *disease*) (also denoted as (*c*, *r*, *d*)).

### 2.2.1 Triples extraction

Following Zhou et al. [17], we extract triples from both the CDR dataset and CTD knowledge base. Triples in CTD are directly extracted. To generate triples of the CDR dataset, we first extract chemical-disease entity pairs. Then, the relations of these pairs are annotated based on CTD. There are three kinds of relations in CTD: *inferred-association*, *therapeutic* and *marker/mechanism*, among which only *marker/mechanism* refers to the true CID relation. For the entity pairs in the CDR data set but not found in CTD, we artificially annotate them with a special relation *null*. Finally, 1,787,913 triples with four relations are obtained for knowledge representation learning.

### 2.2.2 Knowledge representation learning with TransE

TransE [20] is employed to learn knowledge representations in this paper for its simplicity and good performance. All the triples extracted from the CDR dataset and CTD knowledge base are used as correct triples to learn chemical embeddings $\mathbf{e}_c$, disease embeddings $\mathbf{e}_d$ and relation embeddings $\mathbf{r}$ in the common space $\mathbb{R}^k$. TransE models relations as translations from chemicals to diseases, i.e. $\mathbf{e}_c + \mathbf{r} \approx \mathbf{e}_d$ when (*c*, *r*, *d*) holds. The loss function of TransE is defined as follows:

$$L = \sum_{(e_c,r,e_d)\in S} \sum_{\substack{(e'_c,r,e_d) or \\ (e_c,r,e'_d)\in S'}} \max(0, \gamma + \|\mathbf{e}_c + \mathbf{r} - \mathbf{e}_d\| - \|\mathbf{e}'_c + \mathbf{r} - \mathbf{e}'_d\|) \quad (1)$$

where $S$ is the set of correct triples, $S'$ is the set of negative triples, and $\gamma > 0$ is a margin between correct triples and negative triples. The set of correct triples $S$ is extracted from the CDR dataset and CTD knowledge base. The set of negative triples $S'$, according to Formula (1), is constructed with either the chemical or disease in correct triples replaced by a random entity [19].

To get knowledge representations and word embeddings in the common space, we initialize entity embeddings with the average embeddings of entity mention words. Relation embeddings are randomly initialized with the uniform distribution in $[-0.25, 0.25]$. Word2Vec[2] [30] is employed to pre-train word embeddings on the PubMed articles provided by Wei et al. [31].

## 2.3 Relation extraction

Both entity embeddings and relation embeddings are used to capture the important context features related to the focused entity pairs. Figure 2 shows the framework of KCN: two convolutional networks are adopted to capture the context information related to chemicals and diseases, respectively. Each convolutional network is composed of four layers: (1) the embedding layer; (2) the entity-based gated convolutional layer; (3) the relation-based attention pooling layer; (4) the softmax layer.

### 2.3.1 Embedding layer

The input sequences of the two convolutional networks are the same. Given an input SDP sequence $w = \{w_1, w_2, ..., w_n\}$ of a candidate instance, we map each token $w_i$ to a *d*-dimensional embedding $x_i \in \mathbb{R}^d$ to obtain a token embedding sequence $\mathbf{X} = [x_1, x_2, ..., x_n] \in \mathbb{R}^{d \times n}$. Embeddings of dependency relation tags and directions in the sequence are randomly initialized. Similarly, the chemical *c*, disease *d* and relation *r* are also mapped to their embeddings $\mathbf{e}_c \in \mathbb{R}^k$, $\mathbf{e}_d \in \mathbb{R}^k$ and $\mathbf{r} \in \mathbb{R}^k$, respectively.

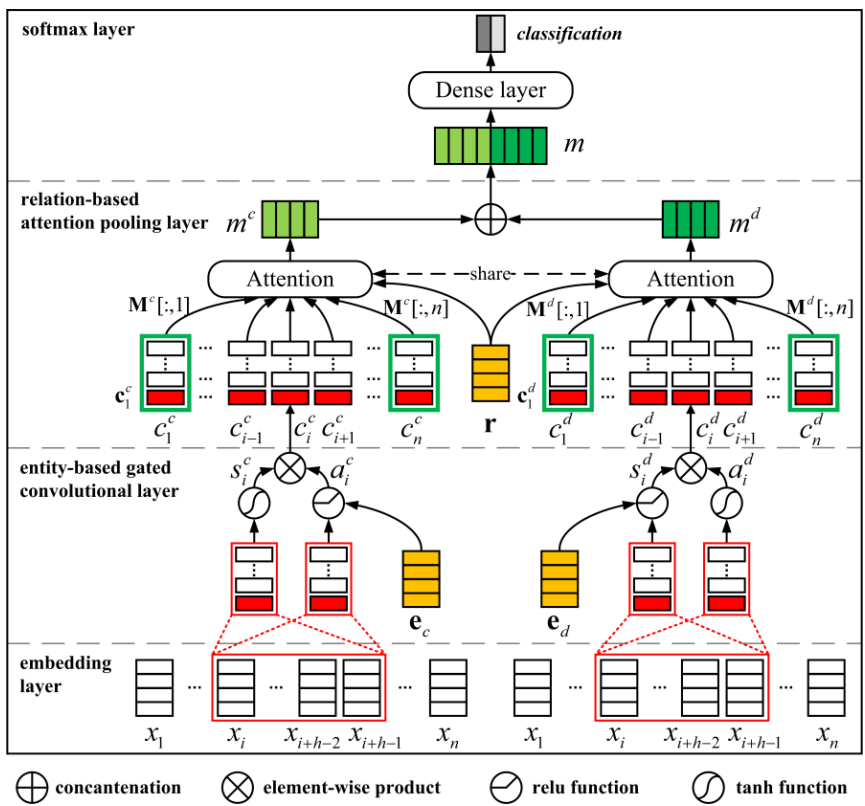

**Fig. 2** The framework of the knowledge-guided convolutional networks.

### 2.3.2 Entity-based gated convolutional layer

Entity-based gated convolutions can selectively extract entity-specific convolutional features with the given entities. Entity-based gated convolutions in the two convolutional networks are performed based on chemical entities and disease entities, respectively.

To help better understand gated convolutions, we first provide a brief review of traditional convolutions. Traditional convolutions apply multiple filters with different widths to get *n*-gram features [32]. Formally, given the input embedding sequence $\mathbf{X}$, the convolution operation at position *i* can be formed as follows:

$$c_i = f(\mathbf{X}_{i:i+h-1} * \mathbf{W}_c + b_c) \qquad (2)$$

where $\mathbf{W}_c \in \mathbb{R}^{d \times h}$ is the filter matrix, *f* is a non-linear activation function, $*$ denotes the convolution operation and $\mathbf{X}_{i:i+h-1}$ refers to the concatenation of *h* token embeddings. The convolution operation maps *h* tokens in the receptive field to a feature $c_i$. Each filter is used for each possible window of *h* tokens in the sequence $\mathbf{X}$ to produce a feature map $\mathbf{c} = [c_1, c_2, ..., c_{n-h+1}] \in \mathbb{R}^{n-h+1}$. If there are *l* filters of the same width *h*, the convolutional features form a matrix $\mathbf{C} = [\mathbf{c}_1, \mathbf{c}_2, ..., \mathbf{c}_l]^T \in \mathbb{R}^{l \times (n-h+1)}$.

Our gated convolutions control the propagation of convolutional features with additional gating units. Inspired by Xue and Li [27], Gated Tanh-ReLU Units (GTRU) are used to control the path through which information flows towards the subsequent pooling layer. GTRU have two nonlinear gates, Tanh and ReLU, each of which is connected to a convolution operation. With entity embeddings, they can selectively ouput the entity-specific convolutional features for CDR extraction.

In the gated convolutional layer, two GTRUs of the same structure are applied to the two entities, respectively. Take the GTRU with chemical embeddings $\mathbf{e}_c$ for illustration. For a token embedding sequence $\mathbf{X} = [x_1, x_2, ..., x_n]$, the convolutional features $c_i$ at position *i* are calculated as follows:

$$s_i^c = \tanh(\mathbf{X}_{i:i+h-1} * \mathbf{W}_s^c + b_s^c)$$

$$a_i^c = \text{relu}(\mathbf{X}_{i:i+h-1} * \mathbf{W}_a^c + \mathbf{V}_a^c \mathbf{e}_c + b_a^c)$$

$$c_i^c = s_i^c \times a_i^c \quad (3)$$

where $\mathbf{W}_a^c, \mathbf{W}_s^c \in \mathbb{R}^{d \times h}$ are the convolution filters of size $h$, $\mathbf{V}_a^c \in \mathbb{R}^{1 \times k}$ is a transition matrix and $b_a^c, b_s^c \in \mathbb{R}^1$ are the biases. The convolution operations for generating convolutional features $a_i^c$ and $s_i^c$ in Formula (3) are the same as traditional convolutions. The convolutional feature $s_i^c$ is only responsible for representing context features. But the convolutional feature $a_i^c$ receives additional chemical embeddings $\mathbf{e}_c$. $a_i^c$ is used to control context features $s_i^c$ to obtain the features $c_i^c$.

This paper uses $l$ filters to obtain the chemical-based context features $\mathbf{M}^c = [\mathbf{c}_1^c, \mathbf{c}_2^c, ..., \mathbf{c}_l^c]^T \in \mathbb{R}^{l \times n}$. Similar to $\mathbf{M}^c$, the disease-based features $\mathbf{M}^d$ are generated through the same gated convolution operations with disease embeddings $\mathbf{e}_d$. The $i$-th column of $\mathbf{M}^c$ ($\mathbf{M}^d$) is defined as a chemical-based (disease-based) context feature vector $\mathbf{M}^c[:,i]$ ($\mathbf{M}^d[:,i]$) as shown in the green boxes in Fig. 2. In fact, $\mathbf{M}^c[:,i]$ ($\mathbf{M}^d[:,i]$) can be seen as the chemical-based (disease-based) context features of the $i$-th token $x_i$.

### 2.3.3 Relation-based attention pooling layer

In traditional CNN, the feature maps generated by the convolutional layer are fed to a max pooling layer to get the most salient features. However, the CDR extraction model should pay more attention to the important context clues of relations between entities.

Following this intuition, the attention mechanism is employed to learn the importance of each entity-based context feature with regard to relation embeddings. In attention pooling layer, the two convolutional networks share the same attention parameters to learn the weights of chemical-based context vectors and disease-based context vectors. Sharing parameters enables the two entities to communicate with each other.

Take the chemical-based context features $\mathbf{M}^c$ as example. For each context vector $\mathbf{M}^c[:,i]$, we use an attention mechanism to compute its semantic relevance with relation embedding $\mathbf{r}$ of the focused entity pair as follows:

$$g_i = \tanh(\mathbf{W}_g \mathbf{M}^c[:,i] + b_g) \odot \mathbf{r} \quad (4)$$

where $\odot$ denotes the dot product, $\mathbf{W}_g \in \mathbb{R}^{k \times l}$ is the transition matrix and $b_g \in \mathbb{R}^k$ is the bias.

After obtaining $\{g_1, g_2, ..., g_n\}$, the attention weight of each context vector can be defined with a softmax function as follows:

$$\alpha_i = \frac{\exp(g_i)}{\sum_{j=1}^{n} \exp(g_j)} \quad (5)$$

Then the weighted sum feature $m^c \in \mathbb{R}^l$ is defined as follows:

$$m^c = \sum_{i=1}^{n} \alpha_i \mathbf{M}^c[:,i] \quad (6)$$

Finally, the two weighted sum entity-based features are concatenated to form the weighted context feature $m = m^c \oplus m^d$.

### 2.3.4 Softmax layer

For the relation classification, a softmax layer is employed on the weighted context feature $m$. It takes feature $m$ as its input and outputs the probability distribution of relation labels. Formally, the softmax layer is defined as follows:

$$o = \text{relu}(\mathbf{W}_h m + b_h)$$

$$p(y = j | T) = \text{softmax}(\mathbf{W}_o o + b_o) \quad (7)$$

where $\mathbf{W}_h \in \mathbb{R}^{h_0 \times 2l}$ and $\mathbf{W}_o \in \mathbb{R}^{2 \times h_0}$ represent the transition matrices, $b_h \in \mathbb{R}^{h_0}$ and $b_o \in \mathbb{R}^2$ are their corresponding biases and $T$ denotes all the training instances.

The cross-entropy loss function is used as the training objective. For each predicted instance $T^{(t)}$ and its golden label $y^{(t)}$, the loss function is defined as follows:

$$loss = -\frac{1}{N}\sum_{t=1}^{N}\log p\left(y^{(t)} \mid T^{(t)}\right) \qquad (8)$$

where $N$ is the number of all the training instances and the superscript $t$ indicates the $t$-th labeled instance.

## 2.4 Relation merging

After the relation extraction at intra- and inter-sentence levels, two sets of prediction results are obtained. We merge them together as the final document level results. Since we extract all the possible candidate instances at intra-sentence level, there might be multi-instances for one entity pair but with inconsistent predictions. In this case, we believe that an entity pair has a CID relation as long as there is at least one instance predicted to be positive.

## 3 Experiments and Results

### 3.1 Experiment setup

#### 3.1.1 Dataset

Experiments are conducted on the BioCreative V Track 3 CDR extraction dataset, which contains a total of 1500 PubMed articles: 500 each for the training, development and test set. The chemicals, diseases and relations are manually annotated with their MeSH IDs [28] and positions in documents. Table 1 describes the statistic of the dataset.

**Table 1** Statistics of the CDR dataset

| Dataset | Articles | Chemical | | Disease | | CID |
|---|---|---|---|---|---|---|
| | | Men | ID | Men | ID | |
| Training | 500 | 5203 | 1467 | 4182 | 1965 | 1038 |
| Development | 500 | 5347 | 1507 | 4244 | 1865 | 1012 |
| Test | 500 | 5385 | 1435 | 4424 | 1988 | 1066 |

Men, ID and CID denotes the number of Mentions, MeSH IDs and CID relations, respectively.

Following Zhou et al. [17], we combine the original training set and development set as the training set: 80% is used for training and 20% for validation. The evaluation is reported by the official evaluation toolkit[3], which adopts Precision (P), Recall (R) and F1-score (F) as the metrics.

#### 3.1.2 Training details

This section describes the training details about the experiments. For knowledge representation learning, we directly run the TransE code[4] released by Lin et al. [22] with 500 epochs. The dimensions of token, entity and relation embeddings are all set to 100. For KCN training, 100 filters with window size $h=1,2,3,4,5$ respectively are used in the gated convolutional layer. We use a batch size of 20 and the Adam optimizer [33] with learning rate: $\lambda_1=0.0001$ at intra-sentence level, $\lambda_2=0.0002$ at inter-sentence level. Table 2 lists the hyper-parameters of KCN.

Our model is implemented with an open-source deep learning framework PyTorch and is publicly available online.

**Table 2** Settings of hyper-parameters.

| Parameter | Description | Value |
|---|---|---|
| $n_k$ | TransE epochs | 500 |
| $d$ | Word embedding dimension | 100 |
| $k$ | Entity/relation embedding dimension | 100 |
| $l$ | Filter number | 100 |
| Mini-batch | Minimal batch size | 20 |
| $\lambda_1$ | Learning rate of intra-sentence instances | 0.0001 |
| $\lambda_2$ | Learning rate of inter-sentence instances | 0.0002 |

### 3.2 Results

#### 3.2.1 Effects of prior knowledge

To investigate the effects of prior knowledge, we compare our **KCN** with its three variants:

**AE (Averaged Entity Embedding)**: This variant represents an entity embedding as the average of its constituting word embeddings. That is to say, only relation embeddings learned from KBs are employed, while entity embeddings learned from KBs are not used.

**SA (Self-Attention)**: This variant replaces the relation-based attention mechanism with a self-attention mechanism, which can be represented as: $g_i = \tanh(\mathbf{w}_g^T \mathbf{M}^c[:,i] + b_g)$. That is to say, only entity embeddings learned from KBs are employed, while relation embeddings learned from KBs are not used.

**AE-SA (Averaged Entity Embedding and Self-Attention)**: This variant represents an entity embedding as the average of its constituting word embeddings, and replaces the relation-based attention mechanism with a self-attention mechanism at the same time. That is to say, neither entity embeddings nor relation embeddings learned from KBs are used.

Table 3 compares **KCN** with the three variants at both intra- and inter-sentence levels. From the table, we can see that:

(1) Compared with **KCN**, **AE** replaces the entity embedding with its corresponding word embeddings and causes the document level F1-score to drop by 2.91%. This indicates that prior knowledge encoded entity embeddings are more effective than entity embeddings expressed by word embeddings.

(2) **SA** discards relation embeddings in **KCN** and causes the F1-score significantly decreases by 12.03%. This suggests that relation embeddings learned from KBs are the direct evidence for CDR extraction.

(3) **AE-SA** achieves the worst results among the three variants. It does not leverage any knowledge representations learned from KBs, resulting in a 13.21% decrease of F1-score.

(4) With the help of the deep semantic relevance between entity embeddings and relation embeddings, **KCN** achieves the highest document level F1-score of 71.28%.

### 3.2.2 Influences of *curated CDR articles*

CTD provides prior knowledge for relation extraction in the CDR dataset. One may then wonder if there is any relation between the curated data in CTD and the CDR dataset. To clarify the doubt, we make a statistic on the CDR dataset and find that all the 1500 articles in the CDR dataset have been curated in CTD. We call these articles as *curated CDR articles*.

To explore the influences of *curated CDR articles*, we remove some triples in *curated CDR articles* (defined as *CDR triples*) from CTD. Three new models are trained based on **KCN**, namely **-train&test**, **-train** and **-test**.

(1) **-train&test** indicates all *CDR triples* in the whole CDR dataset are removed from CTD.

(2) **-train** indicates *CDR triples* in the CDR training and development set are removed from CTD.

(3) **-test** indicates *CDR triples* in the CDR test set are removed from CTD.

From the results shown in Table 4, we can see that:

(1) Without the guidance of *CDR triples* in the CDR dataset, the F1-score drops from 71.28% (**KCN**) to 61.35% (**-train&test**). Once *CDR triples* are removed from CTD, entity pairs in the CDR dataset will be incorrectly annotated as the *null* relation. As a result, they may be misclassified.

(2) Similar to **-train&test**, **-train** and **-test** also make some declines in the document level F1-score.

Based on the experiments above, one may doubt if **KCN** only relies on prior knowledge extracted from CTD. To clarify this, we design an extra model called **Only KB**. This model extracts CID relations by matching the entity pairs in the CDR dataset with the triples in CTD. The results are shown in the last row of Table 4.

(1) Compared with **KCN**, **Only KB** gets a lower F1-score of 63.90%, which demonstrates the importance of the contexts.

(2) **Only KB** has a fairly low precision. CTD curates a large number of CID triples, however, some of which are not annotated as CID relations in the CDR test set. In this case, many negative triples will be wrongly classified as positives through matching.

(3) The recall of **Only KB** is not 100%, which is mainly caused by two reasons. Firstly, our heuristic rules for negative instance filtering (see **subsection "Intra- and inter-sentence level instance constructio**n**"**) remove some positive instances. Secondly, although CTD covers all the articles in the CDR dataset, not all positive entity pairs in the CDR dataset are included in it.

As illustrated above, *curated CDR articles* can be helpful for CDR extraction. And the key to achieving the good performance is the combination of prior knowledge and context information.

### 3.2.3 Effects of architecture

To better understand the architecture of **KCN**, we compare it with two variants:

**w/o GTRU**: This variant replaces GTRU with traditional Tanh, i.e. entity-based gated convolutions degenerate to traditional convolutions. Without the control of entity embeddings, the operations in the two convolutional networks are the same. Therefore, only one convolutional network is enough.

**w/o Att**: This variant replaces the relation-based attention pooling with a max pooling.

From the results shown in Table 5, we can observe that:

(1) Without entity-based gated convolutions, the F1-score of **w/o GTRU** decreases from 71.28% to 68.43%. It is probably that entity-based gated convolutions could extract entity-specific contexts for CDR extraction.

(2) When we remove the attention pooling, the performance of **w/o Att** significantly drops. The possible reason is that the relation-based attention mechanism could find important contexts related to relations.

**Table 3** Effects of different prior knowledge on performance on the CDR dataset

| Method | Intra-sentence level | | | Inter-sentence level | | | Document level | | |
| --- | --- | --- | --- | --- | --- | --- | --- | --- | --- |
| | P (%) | R (%) | F (%) | P (%) | R (%) | F (%) | P (%) | R (%) | F (%) |
| **KCN** | 70.61 | **60.41** | **65.12** | 65.37 | 12.57 | **21.09** | 69.65 | **72.98** | **71.28** |
| **AE** | **71.44** | 57.04 | 63.43 | 57.71 | 10.88 | 18.31[††] | 68.82 | 67.92 | 68.37[†] |
| **SA** | 60.99 | 53.38 | 56.93[††] | 40.57 | 8.07 | 13.46[††] | 57.21 | 61.44 | 59.25[††] |
| **AE-SA** | 58.03 | 53.56 | 55.71[††] | 47.69 | 5.82 | 10.37[††] | 56.82 | 59.38 | 58.07[††] |

The descriptions and analysis for Table 3 could be found in **subsection "Effects of prior knowledge"**. The marker [†] and [††] represent $P$-value <0.05 and $P$-value <0.01, respectively, using pairwise $t$-test against **KCN**. The highest scores are highlighted in bold.

**Table 4** Influences of the curated articles in the CDR dataset on the relation extraction results

| Method | Intra-sentence level | | | Inter-sentence level | | | Document level | | |
| --- | --- | --- | --- | --- | --- | --- | --- | --- | --- |
| | P (%) | R (%) | F (%) | P (%) | R (%) | F (%) | P (%) | R (%) | F (%) |
| **KCN** | **70.61** | 60.41 | **65.12** | **65.37** | 12.57 | 21.09 | **69.65** | 72.98 | **71.28** |
| -train&test | 64.44 | 53.38 | 58.39 | 45.41 | 8.35 | 14.10 | 60.98 | 61.73 | 61.35 |
| -train | 64.07 | 60.23 | 62.09 | 54.63 | 11.07 | 18.41 | 62.40 | 71.29 | 66.55 |
| -test | 65.53 | 48.87 | 55.99 | 49.71 | 8.16 | 14.02 | 61.76 | 57.41 | 59.50 |
| **Only KB** | 59.44 | **65.85** | 62.48 | 31.34 | **21.49** | **26.36** | 50.41 | **87.24** | 63.90 |

The descriptions and analysis for Table 4 could be found in **subsection "Influences of the curated articles in the CDR dataset"**. The highest scores are highlighted in bold.

**Table 5** Effects of each component of architecture on performance on the CDR dataset

| Method | Intra-sentence level | | | Inter-sentence level | | | Document level | | |
| --- | --- | --- | --- | --- | --- | --- | --- | --- | --- |
| | P (%) | R (%) | F (%) | P (%) | R (%) | F (%) | P (%) | R (%) | F (%) |
| **KCN** | **70.61** | 60.41 | **65.12** | **65.37** | **12.57** | **21.09** | **69.65** | **72.98** | **71.28** |
| w/o GTRU | 67.71 | **60.60** | 63.96[†] | 60.95 | 9.66 | 16.68[††] | 66.70 | 70.26 | 68.43[††] |
| w/o Att | 63.37 | 52.25 | 57.28[††] | 42.55 | 9.38 | 15.37[††] | 58.98 | 61.63 | 60.28[††] |

The descriptions and analysis for Table 5 could be found in **subsection "Effects of architecture"**. The marker [†] and [††] represent $P$-value <0.05 and $P$-value <0.01, respectively, using pairwise $t$-test against **KCN**. The highest scores are highlighted in bold.

**Table 6** Effects of different parameter sharing strategies on performance on the CDR dataset

| Method | Intra-sentence level | | | Inter-sentence level | | | Document level | | |
|---|---|---|---|---|---|---|---|---|---|
| | P (%) | R (%) | F (%) | P (%) | R (%) | F (%) | P (%) | R (%) | F (%) |
| **KCN** | **70.61** | 60.41 | **65.12** | 65.37 | **12.57** | **21.09** | 69.65 | **72.98** | **71.28** |
| **SGate-SAtt** | 70.09 | 60.23 | 64.78 | 62.96 | 9.57 | 16.61[††] | 69.02 | 69.79 | 69.40[†] |
| **DGate-DAtt** | 70.39 | 60.23 | 64.91[†] | **68.13** | 10.23 | 17.78[††] | **70.06** | 70.45 | 70.25[††] |
| **SGate-DAtt** | 69.15 | **60.98** | 64.81[†] | 63.19 | 10.79 | 18.43[††] | 68.18 | 71.76 | 69.93[††] |

The descriptions and analysis for Table 6 could be found in **subsection "Effects of sharing parameters"**. The marker [†] and [††] represent *P*-value <0.05 and *P*-value <0.01, respectively, using pairwise *t*-test against **KCN**. The highest scores are highlighted in bold.

**Table 7** Effects of different gating mechanisms in the gated convolutional layer on performance on the CDR dataset

| Method | Intra-sentence level | | | Inter-sentence level | | | Document level | | |
|---|---|---|---|---|---|---|---|---|---|
| | P (%) | R (%) | F (%) | P (%) | R (%) | F (%) | P (%) | R (%) | F (%) |
| **GTRU (KCN)** | 70.61 | **60.41** | **65.12** | **65.37** | **12.57** | **21.09** | 69.65 | **72.98** | **71.28** |
| **GTU** | **71.74** | 59.29 | 64.92[††] | 62.05 | 11.35 | 19.19[††] | **69.98** | 70.64 | 70.31[††] |
| **GLU** | 71.38 | 59.66 | 65.00[††] | 60.11 | 10.32 | 17.61[††] | 69.46 | 69.98 | 69.72[††] |

The descriptions and analysis for Table 7 could be found in **subsection "Effects of gating mechanisms"**. The marker [†] and [††] represent *P*-value <0.05 and *P*-value <0.01, respectively, using pairwise *t*-test against **KCN**. The highest scores are highlighted in bold.

### 3.2.4 Effects of sharing parameters

In **KCN**, the two convolutional networks use different sets of parameters in the gated convolutions but share the same parameters in the attention pooling. To explore the effects of sharing parameters, we compare **KCN** with three variants:

**SGate-SAtt**: In this variant, the parameters in the gated convolutions and the attention pooling are both shared.

**DGate-DAtt**: In this variant, neither the parameters in the gated convolutions nor the parameters in the attention pooling are shared.

**SGate-DAtt**: In this variant, the parameters in the gated convolutions are shared, while the parameters in the attention pooling are not.

From the results shown in Table 6, we can find that:

(1) Compared with **KCN**, **SGate-SAtt** ignores specific information related to each entity, resulting in performance decline.

(2) **DGate-DAtt** focuses on more specific information related to each entity but ignores the connection between the two entities, which leads to a slight drop in the performance.

(3) **SGate-DAtt** captures specific information related to each entity in the attention pooling. The F1-score of **SGate-DAtt** is slightly better than that of **SGate-SAtt**. This demonstrates that entity-specific information is needed for CDR extraction, either in the gated convolutions or in the attention pooling.

### 3.2.5 Effects of gating units

This subsection compares the effects of the different gating units used in the gated convolutions, including **GTRU** [27] (namely **KCN**), Gated Tanh Units (**GTU**) $\tanh(\mathbf{X} * \mathbf{W}_s + b_s) \times \sigma(\mathbf{X} * \mathbf{W}_a + \mathbf{V}_a \mathbf{e} + b_a)$ [26] and Gated Linear Units (**GLU**) $(\mathbf{X} * \mathbf{W}_s + b_s) \times \sigma(\mathbf{X} * \mathbf{W}_a + \mathbf{V}_a \mathbf{e} + b_a)$ [25]. **GTU** and **GLU** have shown their effectiveness in language modeling [25, 26].

Table 7 demonstrates that **GTRU** outperforms the other two gating units. **GTU** and **GLU** use sigmoid gates, whose upper bounds are +1. However, ReLU gates used in **GTRU** have no restrictions on the upper bound. It can amplify knowledge-related context features according to the relevance between context features and entity embeddings.

## 4 Discussion

### 4.1 Visualizations

To illustrate the guidance capacity of prior knowledge in **KCN**, we visualize the weights generated by attention mechanisms and gates in the form of heat maps in Fig. 3 and 4 respectively.

#### 4.1.1 Attention visualization

The attention weights in **KCN** and **AE-SA** are visualized in Fig. 3a and 3b, respectively. Each subfigure has two rows, which correspond to the attention weight of the chemical-based features $\mathbf{M}^c$ and the disease-based features $\mathbf{M}^d$, respectively.

In Fig. 3, the sequence "***fludrocortisone*** ↑ *pmod* ↑ *by* ↑ *vmod* ↑ *reversed* ↑ *vmod* ↑ *induced* ↑ *nmod* ↑ ***hyperkalemia***" is a negative instance for the focused entity pair "*fludrocortisone*" and "*hyperkalemia*". It is correctly classified by **KCN** but misclassified by **AE-SA**.

As can be seen from Fig. 3a, **KCN** pays more attention to the negation word "*reverse*", which helps classify the negative instance correctly. Moreover, the two entities pay attention to each other in Fig. 3a. The relation-based attention could build the links between them.

However, in Fig. 3b, the weights of all the tokens in **AE-SA** have no obvious difference. This may be caused by the lack of prior knowledge. Without its guidance, the attention in **AE-SA** fails to catch the crucial information, resulting in misclassification.

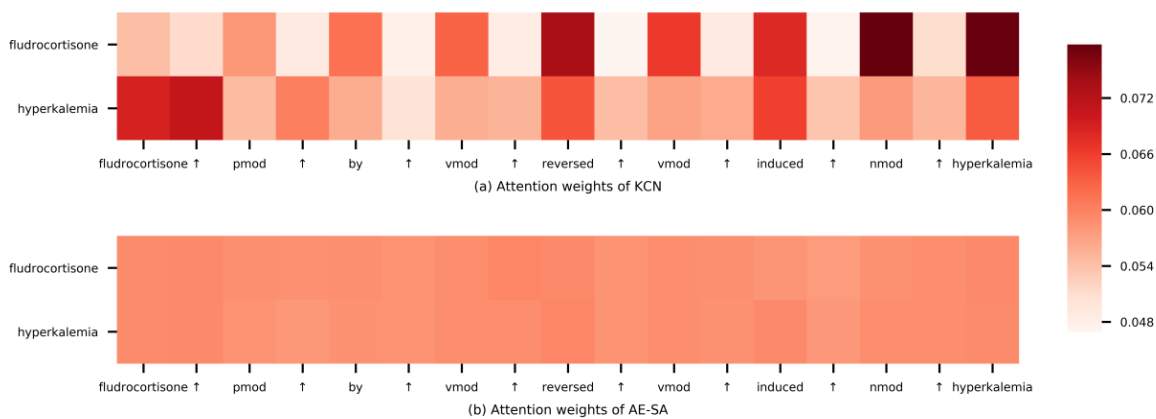

**Fig. 3** The attention visualization of a negative instance.

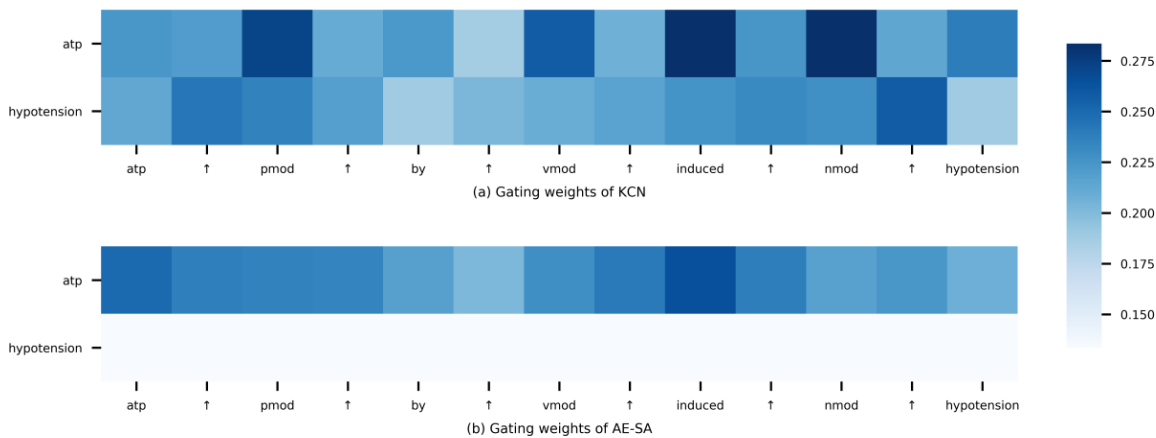

**Fig. 4** The gating visualization of a positive instance.

### 4.1.2 Gating visualization

The weights generated by gates in **KCN** and **AE-SA** are visualized in Fig. 4a and 4b, respectively. For a sequence, there are $n_{token} \times n_{filter} \times n_{dimension}$ outputs of the ReLU gates. We average $n_{filter} \times n_{dimension}$ gate ouputs as the weight of each token. We take a positive instance "***atp*** ↑ *pmod* ↑ *by* ↑ *vmod* ↑ *induced* ↑ *nmod* ↑ ***hypotension***" in Fig. 4 as an example, which is also correctly classified by **KCN** but misclassified by **AE-SA**.

As can be seen from Fig. 4a, with the guidance of prior knowledge, the chemical "***atp***" controlled gates assign more weights on the trigger word "*induced*", which is an important cue for positive instance classification. However, in Fig. 4b, each token weight controlled by disease "***hypotension***" drops dramatically. Due to the loss of the crucial cue, the instance is misclassified as negative by **AE-SA**.

### 4.2 Comparison with related works

#### 4.2.1 Comparison with previous systems

We compare **KCN** with previous systems of the BioCreative V CDR Task in Table 8. To make a fair comparison, all the systems are evaluated on the CDR test set with the golden standard entity annotations. The systems can be divided into 2 groups: systems without KBs and systems with KBs.

**Table 8** Comparison with previous systems of CDR extraction

| Method | | System | P (%) | R (%) | F (%) |
|---|---|---|---|---|---|
| without KBs | Feature-based | Gu et al. [6] | 62.00 | 55.10 | 58.30 |
| | Neural network-based | Nguyen et al. [13] | 57.00 | 68.60 | 62.30 |
| | | Le et al. [14] | 58.02 | 76.20 | **65.88** |
| | | Verga et al. [15] | 55.60 | 70.80 | 62.10 |
| with KBs | Feature-based | Pons et al. [9] | 73.10 | 67.60 | 70.20 |
| | | Peng et al. [10] | 68.15 | 66.04 | 67.08 |
| | | ♣Peng et al. [10] | 71.07 | 72.61 | **71.83** |
| | Neural network-based | Li et al. [16] | 59.97 | 81.49 | 69.09 |
| | | Zhou et al. [17] | 60.51 | 80.48 | 69.08 |
| | | **Ours** | 69.65 | 72.98 | **71.28** |
| | | ♣**Ours** | 72.12 | 68.67 | 70.35 |

The descriptions and analysis for Table 8 could be found in **subsection "Comparison with previous works"**. The marker ♣ indicates that the system uses additional weakly labeled data for training. The highest F1-score of each subgroup is highlighted in bold.

From Table 8, we can see that systems with KBs outperform systems without KBs. This indicates that prior knowledge can be an effective promotion for CDR extraction.

Among the systems without KBs, neural network-based methods [13-15] perform better than feature-based methods [6], which shows the strength of low-dimensional feature vectors in context modeling. Particularly, Le et al. [14] employ the SDP between chemical and disease entities with a CNN-based model, and achieve the highest F1-score of 65.88% among them. However, their system lacks the guidance of prior knowledge. Only using the context information limits the performance of their system.

As for the systems with KBs, Peng et al. [10] use support vector machines (SVM) with one-hot knowledge features extracted from CTD and achieve an F1-score of 67.08%. Furthermore, ♣Peng et al. [10] introduce additional weakly labeled data to improve the F1-score to 71.83% (4.75% increase). Inspired by ♣Peng et al. [10], we also add the same weakly labeled data to train our **KCN**. However, the document level F1-score slightly drops to 70.35%, which could be found in the last row of Table 8 (♣**Ours**).

Weakly labeled data often contains some noise, which may harm the system performance. Generally, an effective method of reducing noise is usually needed [34]. However, without any denoising mechanism, ♣Peng et al. [10] still show strong performance boost from using this data. The good anti-noise capacity may benefit from the careful feature engineering and shallow learning methods. Different from ♣Peng et al. [10], **KCN** uses low-dimensional features with deep learning methods. If an effective denoising mechanism is not applied, **KCN** seems unable to deal with noise.

Pons et al. [9] leverage rich one-hot knowledge features extracted from a commercial system Euretos Knowledge Platform. Different from them, our system represents knowledge with low-dimensional embeddings rather than one-hot features. This could create the correlations between features in a low-dimensional vector space.

Li et al. [16] map knowledge features into low-dimensional vectors to help CDR extraction. Zhou et al. [17] use TransE to learn knowledge representations and incorporate relation embeddings with contexts through an attention mechanism. Different from them: 1) we introduce both entity and relation embeddings, while Zhou et al. [17] only take relation embeddings into consideration; 2) we use both the gating units and the attention mechanism to select the important context features, while Zhou et al. [17] only use the attention mechanism. Therefore, we achieve a better performance than Zhou et al. [17].

### 4.2.2 Influences of *CDR triples* on previous systems

To further explore the influences of *CDR triples* on previous systems, we reproduce two representative systems, named **Zhou-feature** and **Zhou-CAN**, to represent feature-based methods and neural network-based methods, respectively. **Zhou-feature** is the polynomial kernel-based system without using prior knowledge in Zhou et al. [12]. **Zhou-CAN** is the CNN-based system with prior knowledge in Zhou et al. [17].

**Table 9** Comparison with previous systems under the three conditions

| Condition | System | P (%) | R (%) | F (%) |
|---|---|---|---|---|
| (I) | ♣**Zhou-feature** | 62.15 | 46.28 | 53.70 |
| | ♣**Zhou-CAN** | 48.24 | 66.89 | 56.05 |
| | ♣**KCN (AE-SA)** | 56.82 | 59.38 | 58.07 |
| (II) | ♥**Zhou-feature** | 68.55 | 59.10 | 63.48 |
| | ♥**Zhou-CAN** | 60.51 | 80.48 | 69.08 |
| | ♥**KCN** | 69.65 | 72.98 | **71.28** |
| (III) | ♦**Zhou-feature** | 59.95 | 45.78 | 51.91 |
| | ♦**Zhou-CAN** | 60.30 | 55.72 | 57.92 |
| | ♦**KCN** | 62.68 | 57.04 | 59.73 |

The descriptions and analysis for Table 9 could be found in **subsection "Influences of *CDR triples* on previous works"**. The three different conditions are (I) without KBs, marked as ♣; (II) with KBs, marked as ♥; (III) with KBs but removing *CDR triples* in the CDR test set, marked as ♦. The highest F1-score is highlighted in bold.

Table 9 compares **KCN** with **Zhou-feature** and **Zhou-CAN** under the following three conditions: (I) without KBs, (II) with KBs, (III) with KBs but removing *CDR triples* in the CDR test set. From the results, we can see that:

(1) When *CDR triples* are removed, there is a sharp drop of performance for all the systems, which demonstrates that *CDR triples* play an important role in both feature-based methods and neural network-based methods.

(2) Comparing the condition (I) with the condition (III), the performance of neural network-based methods is improved, while the feature-based method performance is decreased. Once *CDR triples* are removed, relation features extracted from KBs are incorrect. Even though, neural network-based methods could still capture the potential semantic information through context features to remedy the incorrect relations, while feature-based methods could not.

(3) For the two neural network-based methods, ♦**KCN** performs better than ♦**Zhou-CAN** under the condition (III). The reason is probably that both entity and relation embeddings are used in ♦**KCN**, while only relation embeddings are employed in ♦**Zhou-CAN**. Entity embeddings could still give ♦**KCN** effective guidance for CDR extraction.

### 4.3 Error analysis

We perform an error analysis on the final results of KCN to detect the origins of false positives (**FPs**) and false negatives (**FNs**). Figure 5 depicts the distribution of all the errors.

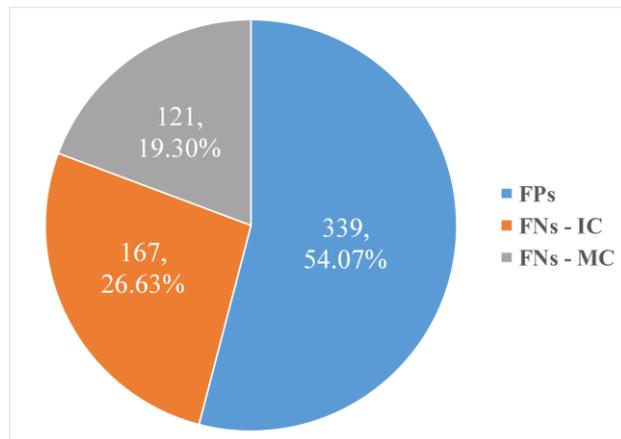

**Fig. 5** The error distribution of origins of **FPs** and **FNs**. **FNs-IC** denotes incorrectly classified **FNs** and **FNs-MC** denotes missing classified **FNs**.

For **FPs** in Fig. 5, 339 negative chemical-disease entity pairs are wrongly classified as positive by KCN, accounting for 54.07% of total. This may be caused by the following two reasons. First, CTD curates a large number of entity pairs with CID relations. But some of them are not annotated as CID relations in the CDR test set, which will mislead their classification. Second, the complex contexts would make it difficult for KCN to distinguish whether the entity pairs have CID relations. Take sentence 2 as an example.

Sentence 2: *The homozygous Gunn rats have unconjugated* **hyperbilirubinemia** *due to the absence of glucuronyl transferase, leading to marked* **bilirubin** *deposition in renal medulla and papilla.*

In this sentence, chemical "*bilirubin*" and disease "*hyperbilirubinemia*" have no CID relation. However, the context surrounding this entity pair is quite confusing, with the phrases "*due to*" and "*leading to*" expressing the meaning of "inducing". This causes the wrong classification.

For **FNs**, there are two main error types:

(1) **FNs-Incorrect Classification (FNs-IC)**: The **FNs-IC** type brings 167 errors with a proportion of 26.63%. Take sentence 3 as an example.

Sentence 3: *BACKGROUND: Several studies have demonstrated liposomal* **doxorubicin** *(Doxil) to be an active antineoplastic agent in platinum-resistant ovarian cancer, with dose limiting toxicity of the standard dosing regimen (50 mg/m(2) q 4 weeks) being severe erythrodysesthesia ("***hand-foot syndrome***") and stomatitis.*

In Sentence 3, chemical "**doxorubicin**" and disease "**hand-foot syndrome**" have a CID relation. However, KCN misclassifies it as negative, which may be caused by the complex contexts, obscure semantic expression and lack of trigger words.

(2) **FNs-Missing Classification (FNs-MC)**: Such error type is due to some positive instances being removed by the heuristic rules mentioned in **subsection "Intra- and inter-sentence level instance construction"**. And it results in 121 errors with a proportion of 19.30%.

## 5 Conclusions

This paper proposes a novel CDR extraction model KCN, which includes the entity-based gated convolutions and relation-based attention pooling. The gating units are employed to control the propagation of context features toward a chemical-disease pair. The attention mechanism is used to focus contexts on the CID relation. The experimental results on the BioCreative V CDR dataset show that KCN could effectively integrate prior knowledge and contexts for the performance improvement.

As future work, we would like to introduce weakly labeled data and consider how to utilize effective denoising mechanisms to purify them.

### Endnotes

[1] http://ctdbase.org/

[2] http://code.google.com/p/word2vec/

[3] http://biocreative.org/tasks/biocreative-v/track3-cdr/

[4] https://github.com/thunlp/KB2E/